\documentclass[10pt,journal,compsoc]{IEEEtran}
\usepackage[sectionbib]{chapterbib}
\usepackage{booktabs}
\usepackage{amsthm}
\theoremstyle{plain}
\usepackage[compress]{cite}
\usepackage{graphics}
\usepackage{epstopdf}
\usepackage{epsfig}
\usepackage{lineno,hyperref}
\usepackage{array}
\usepackage{amsmath,amssymb}
\usepackage{mathrsfs}
\usepackage{subfigure} 
\usepackage{multirow}
\usepackage{xcolor}
\usepackage{bm}
\usepackage{eqparbox}
\usepackage{algorithm} 
\usepackage{algorithmic} 
\usepackage{multirow} 
\usepackage{amsmath} 
\usepackage{xcolor}
\usepackage{graphicx}
\newtheorem{myTheo}{Theorem}
\newtheorem{myLemma}{Lemma}
\usepackage{amsfonts}
\usepackage{pifont}
\usepackage{enumerate}
\usepackage{indentfirst}
\usepackage{url}

\usepackage{ragged2e}

\begin{document}
	
	\title{Deep Manifold Learning with Graph Mining}
	\author{Xuelong Li,\IEEEmembership{~Fellow,~IEEE}, Ziheng Jiao, Hongyuan Zhang,   and Rui Zhang$^*$\IEEEmembership{~Member,~IEEE}
		
		\thanks{$*$ Rui Zhang is the Corresponding Author.}
		

		\thanks{Rui Zhang, Ziheng Jiao, Hongyuan Zhang, and Xuelong Li are with School of Computer Science and School of Artificial Intelligence, Optics and Electronics (iOPEN), Northwestern Polytechnical University, Xi'an 710072, Shaanxi, P. R. China.}

		\thanks{E-mail: ruizhang8633@gmail.com; jzh9830@163.com;  hyzhang98@gmail.com; xuelong\_li@ieee.org.}
		
	}
	

	\IEEEtitleabstractindextext{
		\begin{abstract}
			\justifying 	
			Admittedly, Graph Convolution Network (GCN) has achieved excellent results on graph datasets such as social networks, citation networks, etc. 
			However, softmax used as the decision layer in these frameworks is generally optimized with thousands of iterations via gradient descent. Furthermore, due to ignoring the inner distribution of the graph nodes, the decision layer might lead to an unsatisfactory performance in semi-supervised learning with less label support. 
			To address the referred issues, we propose a novel graph deep model with a non-gradient decision layer for graph mining. Firstly, manifold learning is unified with label local-structure preservation to capture the topological information of the nodes. Moreover, owing to the non-gradient property, closed-form solutions is achieved to be employed as the decision layer for GCN. Particularly, a joint optimization method is designed for this graph model, which extremely accelerates the convergence of the model. Finally, extensive experiments show that the proposed model has achieved state-of-the-art performance compared to the current models.
			
			
		\end{abstract}
		
	}
	\maketitle
	
	\IEEEdisplaynontitleabstractindextext
	
	\IEEEpeerreviewmaketitle
	
	\begin{figure*}[t]
		\centering
		\includegraphics[width=180mm]{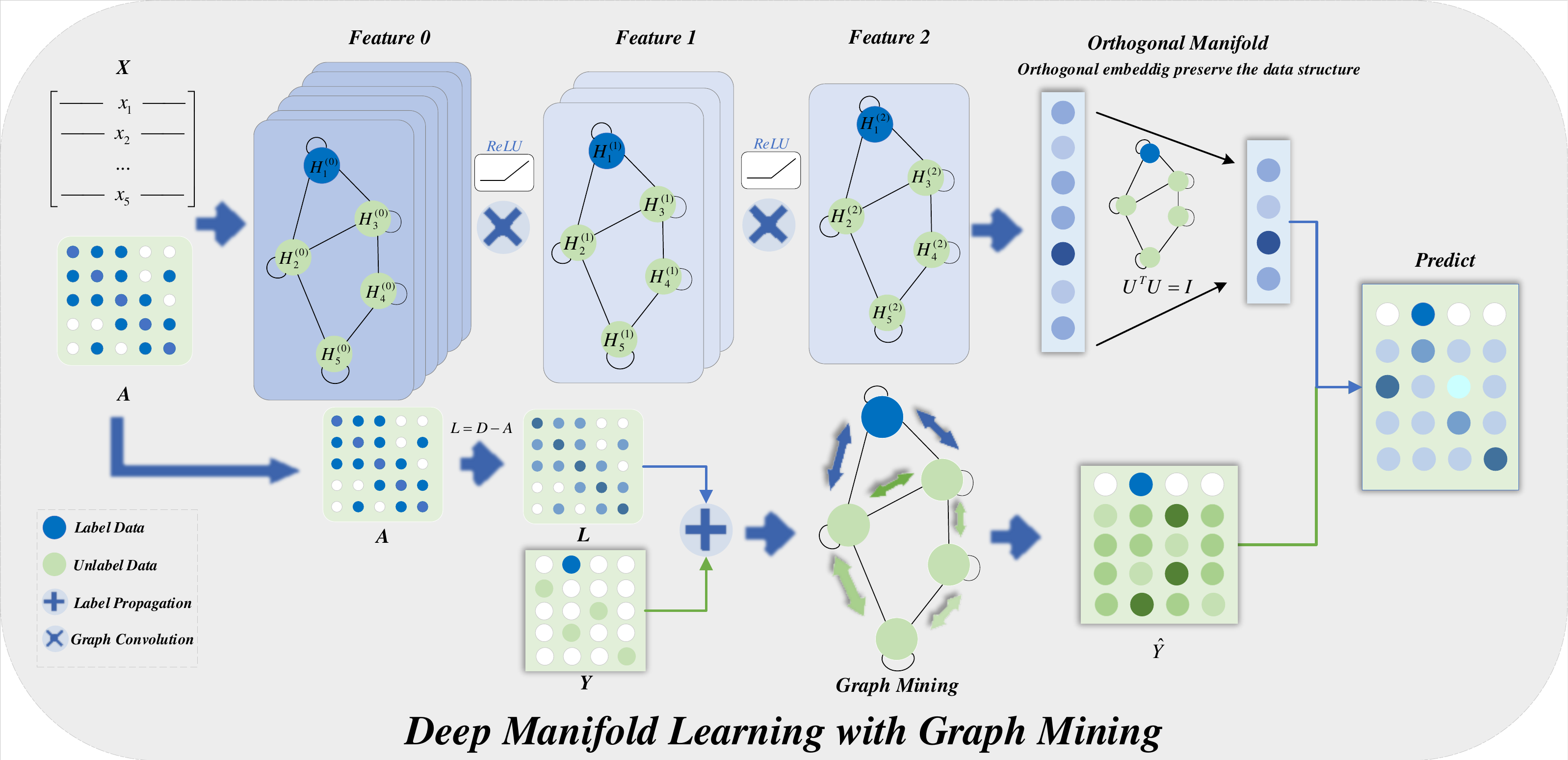}
		\caption{A framework of the proposed model. The dark blue node is the label data and the gray node is the unlabeled data. As is shown, the topology of the prior graph is fully embedded into the model. After the deep graph network extracts the graph embedding, the non-gradient decision layer, unifying the orthogonal manifold and label local-structure preservation, mines the distribution of these embeddings to predict the labels.}
		\label{framework}
		\vspace{-3mm}
	\end{figure*}
	
	\IEEEraisesectionheading{\section{Introduction}\label{sec:introduction}}
	
	\IEEEPARstart{C}{lassification},a classic task in machine learning, aims to classify the unlabeled data by learning the information from the labeled data. However, considering the expensive handcraft labeling, privacy, and security, it is difficult to obtain abundant labeled data for training classifiers. In recent years, semi-supervised algorithms trained with scarce labeled data make some progress in the practical application\cite{zhu2009introduction}. In general, the semi-supervised models could be divided into two categories, based on label information and based on the data distribution\cite{zhu2005semi} .

	Based on label information, these models mainly focus on learning high-quality features from the labeled data, such as co-training method \cite{zhou2005semi, sindhwani2005co, zhou2007semisupervised, mao2009semi}.
	Besides, latent structure information often helps to enhance classification accuracy \cite{2002Learning, long2019locality, iscen2019label, gao2016segmentation, 2018Semi}. Among them, manifold learning is a classical method and can explore the latent topological information of the data \cite{criminisi2012decision}. The KPCA \cite{chatpatanasiri2010unified} integrates manifold learning into semi-supervised dimensionality reduction. Besides, a novel generalized power iteration (GPI) \cite{2017A} is employed to solve the problem on the Stiefel manifold. Accordingly, the orthogonal regression manifold (OR) is the least square problem defined on the Stiefel manifold. 
	
	
	However, it is known that the classic convolution neural network mainly deals with regular data such as images and weakly handles irregular data tasks like graph node classification. Some researchers utilize spectral decomposition to explore the structure of the graph and expand the deep network into irregular data space \cite{2020A, dong2021equivalence,2019Masked,yang2016revisiting }. \cite{2020A} offers flexibility to compose and evaluate different deep networks using classic machine learning algorithms. Besides, the pseudo-labels generated by label propagation are utilizing to train graph convolution networks\cite{dong2021equivalence}. Masked-GCN \cite{2019Masked} only propagates  some attributes of nodes to the neighbors, which jointly considers the distribution of local neighbors. Apart from that, \cite{yang2020rethinking} proposes Propagation-regularization to boost the performance of existing GNN models. Aiming to accelerate the speed of the model, \cite{2020Factorized} suggests a method called distant compatibility estimation.
	
	Generally, softmax utilizes the deep feature to predict the label and is extensively employed as the decision layer in the graph networks. However, it ignores the distribution and latent structure of the irregular data during classification. Apart from that, this decision layer is optimized via gradient descent with thousands of iterations to obtain the approximate solutions. Besides, due to the lack of sufficient prior labels under the semi-supervised tasks, trivial solutions can not be avoided during the optimization.
	
	
	To tackle the referred deficiency, we propose a novel deep model for graph knowledge learning with a non-gradient decision layer. Our major contributions are listed by the following items:
	\begin{itemize}
		\item [1)] Unifying the orthogonal manifold with label local-structure preservation to mine the topological information of the deep embeddings, the novel non-gradient graph decision layer is put forward.
		
		\item [2)] The theorems are designed to solve the proposed layer with an elegant analytical solution, which accelerates the convergence of the model.
		
		\item [3)] A novel deep graph convolution network is proposed and optimized jointly.
		Moreover, extensive experiments suggest that the model has achieved state-of-the-art results.
	\end{itemize}

	\section{Notations and Background}
	
	\subsection{Notations}
	In this paper, the boldface capital letters and boldface lower letters are employed to present matrics and vectors respectively. Suppose the $\bm{m_i}$ as the $i$-th row of the matrix $\bm{M} \in R^{d \times c}$. The element of $\bm{M}$ is $m_{ij}$. Besides, the $F$-norm of the matrix $\bm{M}$ is defined as $||\bm{M}||_F=\sqrt{\sum_{i=1}^d \sum_{j=1}^{c} m_{ij}^2}$. The trace and transpose of matrix $\bm{M}$ are denoted as $Tr(\bm{M})$ and $\bm{M}^T$, respectively. $\nabla_{\bm x} f(\bm x)$ means the the gradient of $f(\bm x)$ w.r.t. $\bm x$. And $\bm{1}_c$ is a unit column vector with dimension c. For the semi-supervised classification, $\bm X = [\bm{x_1, x_2,...,x_n}] \in R^{n \times d}$ is the feature matrix with $n$ samples and $d$ dimensions. And we have $l$ labeled data points and $u$ unlabeled data samples where $l+u=n$. Let $\bm{Y}=[\bm{Y}_{l};\bm{Y}_{u}] \in R^{n \times c}$ denotes the labeled and unlabeled with $c$ classes. And $\bm{Y}_{u}$ is initialized according to $\bm{Y}_{u} \bm{1}_c = \bm{1}_{u}$. Apart from that, we denote $\bm{H}^{(m)} \in R^{n \times d_m}$ as the $m$-th layer hidden feature.
	
	\subsection{Background}
	The classic convolutional neural network is designed to process regular data like images. However, there exists numerous irregular data, such as citation networks and social networks. Owing to the distribution of these data is not shift-invariant, the traditional neural networks could not extract the latent topological information of these data. Aiming to tackle these issues, some researchers try to transfer traditional Laplacian kernel and convolution operators into graph data space by spectral decomposition based on the graph \cite{bruna2013spectral, defferrard2016convolutional, hammond2011wavelets}.  However, the spectral decomposition of the Laplacian matrix is very time-consuming. Defferrard et al \cite{defferrard2016convolutional} employ the Chebyshev polynomial to fit the results of  spectral decomposition and design a Chebyshev convolution kernel as follows
	\begin{equation}
		\label{Chebyshev}
		g_{\bm \theta} = g_{\bm \theta}(\bm \Lambda) \approx \sum_{i=0}^{K} \theta_iT_k(\tilde{\bm \Lambda}),
	\end{equation}
	where $\tilde{\bm \Lambda}=2\bm \Lambda/\lambda_{max}- \bm{I}_N$ is the normalized feature vector matrix, $\lambda_{max}$ is the spectral radius, $\bm \theta \in R^K$ is Chebyshev coefficient vector, and the Chebyshev polynomial is defined as $T_k(x)=2xT_{k-1}(x)-T_{k-2}(x)$.
	
	Based on this, Kipf et al \cite{kipf2016semi} use the first-order expansion of the Chebyshev polynomial to fit the graph convolution kernel with $K=1,\lambda_{max}=2$. To prevent overfitting and reduce complexity, the authors introduce some tricks, $\bm \theta_0 = -\bm \theta_1$, $\bm{\tilde{A}}=\bm A + \bm{I}_N$ and the diagnonal element in degree matrix is $\widetilde{\bm D}_{ii}=\sum_{j}\widetilde{\bm A}_{ij}$. Finally, the classic GCN is defined as follows
	\begin{equation}
		\label{GCN}
		\bm{H}^{(l+1)} = \sigma (\widetilde{\bm D}^{-\frac{1}{2}} \widetilde{\bm A} \widetilde{\bm D}^{-\frac{1}{2}}\bm{H}^{(l)} \bm{W}^{(l)}).
	\end{equation}
	
	
	Apart from that, manifold learning \cite{cayton2005algorithms} is a classic nonlinear algorithm for dimensionality reduction, representation learning, and classification tasks. It believes that the data nodes are generated from a low-dimensional manifold embedded in a high-dimensional space and this procedure is often described by a function with few underlying parameters. Based on this idea, manifold learning aims to uncover the function in order to explore the latent topological structure of the data.
	
	\section{Deep Manifold Learning with Graph Mining}
	In this section, we propose a novel graph deep model with a non-gradient decision layer  for graph mining. The framework is illustrated in Fig. \ref{framework}.
	
	\subsection{Problem Revisited: GCN with Softmax}
	
	As is known to us, Softmax has been widely used as a decision layer in lots of classic neural networks. Moreover, on some computer vision tasks such as image classification, these networks also have made excellent progress and can be generally formulated as
	\begin{equation}
		\label{Neural Network}
		\left\{ 
		\begin{array}{l}
			\bm{H}^{(a)} = \sigma (f_n(\bm{W}^{(a-1)}, \bm{H}^{(a-1)} ))\\
			\bm{\hat{Y}} = softmax(\bm{H}^{(m)})\\
		\end{array} 
		\right.
		,
	\end{equation}
	where $f_n$ is a neural network such as CNN to extract the deep features, $a \in \{0,1,...,m\}$ is the $a$-th layer in the network, and $\bm{W}^{(a)}$ is the weight matrix in the $a$-th layer. The latent feature in the $a$-th layer is defined as $\bm{H}^{(a)}$ and the $\bm{H}^{(0)}$ is equivalent to $\bm{X}$. $\bm{\hat{Y}} \in R^{n \times c}$ is the prediction label. The latent feature $\bm H^{(m)}$ has the same shape with $\bm{\hat{Y}}$. $\sigma (\cdot)$ is the activation function. Aiming to predict the classes, Softmax maps and normalizes these deep features into real numbers in $[0, 1]$ via the formula defined as
	\begin{equation}
		\label{softmax}
		\hat{y}_{ij} = \frac{e^{h^{(m)}_{ij}}}{\sum_{j=1}^{c} e^{h^{(m)}_{ij}}}.
	\end{equation}
	The cross-entropy is employed as the loss function and optimized via gradient descent. Aiming to classify the irregular data such as citation network, \cite{kipf2016semi} defines the graph convolution to extract the graph deep features and also classify them via softmax.
	
	From the Eq. (\ref{softmax}), softmax mainly forecasts the probability according to the value of the embedding. However, it ignores the latent topological distribution of the graph embedding. For example, in social networks, people ought to have a high degree of similarity to those with whom they socialize frequently. Therefore, introducing the softmax into GCN for irregular data classification may not get the same achievements as utilizing CNN for the regular data without this topological structure. More importantly, the softmax is optimized via gradient descent with thousands of iteration to obtain approximate solutions, whose time complexity is non-linear with the number of the convolutional layers. Aiming to improve the performance of the GCN, we propose a graph neural network with a non-gradient decision layer.
	
	\subsection{ Graph-based Deep Model with a Non-Gradient Decision Layer}
	
	\subsubsection{Unify Orthogonal Manifold and Label Local-structure Preservation}
	
	Owing to obtaining the closed-form result via the analytical method, the least square regression can also be a classic method to predict the label in machine learning. Apart from this, the softmax has low complexity and high interpretability. In semi-supervised learning, it can be defined as
	\begin{equation}
		\label{Primal_Regression}
		\min \limits_{\bm{U}, \bm{Y}_{u}} ||\bm{X} \bm{U} - [\bm{Y}_{l};\bm{Y}_{u}]||_F^2,
	\end{equation}
	where $\bm{X} \in R^{n \times d}$ is the feature matrix, $\bm{U} \in R^{d \times c}$ is the projection matrix with $c$ classes, and $[\bm{Y}_{l};\bm{Y}_{u}]=\bm{Y} \in R^{n \times c}$ is a indices matrix including one-hot labeled and unlabeled points. We can easily optimize the projection matrix $\bm{U}$ via taking the derivative of Eq. (\ref{Primal_Regression}) w.r.t $\bm{U}$ and setting it to 0,
	\begin{equation}
		\label{primial_optimize_U}
		\nabla_{\bm U} ||\bm{X} \bm{U} - [\bm{Y}_{l};\bm{Y}_{u}]||_F^2 = 0.
	\end{equation}
	
	Unfortunately, in the semi-supervised or unsupervised task, Eq. (\ref{primial_optimize_U}) may cause the trivial solution. The label rate is greatly small in semi-supervised learning, $i.e.$, $l \ll u$. Under this circumstance, when $\bm U = [\bm{1}_{d}, \bm{0}_{d \times (c-1)}]$, the trivial solution that $\bm{Y}_{u}=[\bm{1}_n,\bm{0}_{n \times (c-1)}]$
	may be triggered. Apart from the potential trivial solution, directly projecting the  feature into the label space may ignore the original data structure.
	
	In order to tackle these problems, we firstly introduce the orthogonal manifold, $i.e.$, $\bm{U}^T \bm{U}=\bm{I}$, to learn the low-dimensional distribution of the graph data, which can avoid projecting the unlabeled graph embedding into the same class and improve the robustness of the model. Meanwhile, since the structural similarity and local information are often credible (especially in manifold learning), a label local-structure preservation is unified into the learned low-dimensional distribution. It guides the classification via label similarity measured by local structure and the decision layer can be defined as
	\begin{equation}
		\label{manifold_lp_loss}
		\min \limits_{\bm{U}^T \bm{U}=I, \bm{Y}_{u}} ||\bm{X} \bm{U} - [\bm{Y}_{l};\bm{Y}_{u}]||_F^2 \\
		+ \lambda \sum_{i,j=1}^n a_{ij}  \|\bm y_i - \bm y_j\|_2^2,
	\end{equation}
	where the element of adjacent matrix $\bm A$ is $a_{ij}$, $\bm y_i$ is the one-hot label in $\bm Y=[\bm{Y}_l;\bm{Y}_u]$, and $\lambda$ is a trade-off parameter. The Eq. \ref{manifold_lp_loss} can be directly solved with the proposed Theorem \ref{cal_U} and Theorem \ref{cal_Y_u}, which is non-gradient and proved in Section \ref{section_opt}.
	
	\begin{myTheo} 
		\label{cal_U}
		Given a feature matrix $\bm X$ and a label matrix $\bm Y$, the problem 
		$\min \limits_{\bm{U}^T \bm{U}=I} ||\bm{X}\bm{U}-\bm{Y}||_F^2$
		can be solved by $\bm{U}=\bm{S}[:,:c]$, where $\bm{S}=\bm{E}\bm F^T$ and $[\bm E, \sim, \bm{F}^T]=svd(\bm X \bm{R})$. $\bm R$ is defined as $\bm{R}=[\bm{Y}, \bm{X}\bm{V}]$ and $\bm V \in R^{d \times (d-c)}$ is generated in the orthogonal complement spaces of $\bm U$.
	\end{myTheo}
	
	\begin{myTheo} 
		\label{cal_Y_u}
		Given a feature matrix $\bm X$, a projection matrix $\bm U$ and the labeled matrix $\bm Y_l $ a label matrix, the problem defined as 
		\begin{equation}
			\label{Y_u_pro}
			\begin{split}
				\min \limits_{\bm{Y}_{u}} ||\bm{X} \bm{U} - [\bm{Y}_{l};\bm{Y}_{u}]||_F^2 
				+ \lambda \sum_{i,j=1}^n a_{ij}  \|\bm y_i - \bm y_j\|_2^2
			\end{split}
		\end{equation}
		can be solved by $\bm{Y}_u=(\bm I + \lambda \bm{L}_{uu})^{-1}(\bm{X}_u\bm{U}-\lambda \bm{L}_{ul}\bm{Y}_l)$, where $\bm X=[\bm X_l;\bm X_u]$, $\bm D=diag(\sum_{j=1} a_{ij})$ and Laplacian
		\begin{equation}
			\bm{L}= \bm{D} - \bm{A}=
			\left[
			\begin{array}{cc}
				\bm{L}_{ll} & \bm{L}_{lu} \\
				\bm{L}_{ul} & \bm{L}_{uu}
			\end{array}
			\right].
		\end{equation}
	\end{myTheo}
	
	\subsubsection{A Novel Graph Deep Network}
	
	Because the orthogonal manifold projection is a linear transform, Eq. (\ref{manifold_lp_loss}) has not good performance when facing non-linear problems. As is known to us, the kernel trick and neural network are widely employed to handle the nonlinearity. However, the kernel trick is weak in representation and severely sensitive to the choice of the kernel function. On the contrary, graph convolution not only has an excellent ability in representation learning for irregular data  but also successfully mines the complex relationships and interdependence between graph nodes via spectral and spatial operators. Based on this, we finally put forward a novel GCN with a non-gradient decision layer like
	\begin{equation}
		\label{proposed model}
		\left\{ 
		\begin{split}
			&\bm{H}^{(a)} = \sigma (\varphi(\widetilde{\bm A}) \bm{H}^{(a-1)} \bm{W}^{(a)} + \bm{b}^{(a)})\\
			&\begin{split}
				\min \limits_{\bm{U}^T \bm{U}=I, \bm{Y}_{u}} &||\bm{H}^{(m)} \bm{U} - [\bm{Y}_{l};\bm{Y}_{u}]||_F^2 \\
				&+ \lambda \sum_{i,j=1}^n a_{ij}  \|\bm y_i - \bm y_j\|_2^2,\\
			\end{split}
		\end{split} 
		\right.
	\end{equation}
	where the self-loop adjacent matrix is $\widetilde{\bm A} = \bm A + \bm I \in R^{n \times n}$, and the graph kernel is $\varphi(\bm A)=\widetilde{\bm D}^{-\frac{1}{2}} \widetilde{\bm A} \widetilde{\bm D}^{-\frac{1}{2}}$. Moreover, the whole network is jointly optimized via Algorithm \ref{the_algorithm_to_GCN-OCLP}.
	
	In this network, the non-gradient decision layer not only preserve and utilize the topological information of the deep graph embeddings but also can be solved with the closed-form solutions, which significantly improves the accuracy and accelerates the convergence.
	
	\noindent \textbf{The Merits of the Proposed Model: } 
	Compared with the classic softmax decision layer, the proposed model unifies the orthogonal manifold and label local-structure preservation to learn the distribution and topology of the graph nodes, which successfully avoids the trivial solutions and improves the robustness. Moreover, contrasting with optimized via gradient descent with thousands of iterations, our model can be solved with the analytical solutions via the proposed non-gradient theorems. Finally, we propose a novel deep graph convolution network and design a joint optimization strategy.
	
	\begin{algorithm}[t]
		\caption{Deep Graph Model with a Non-Gradient Decision Layer}
		\label{the_algorithm_to_GCN-OCLP}
		\begin{algorithmic}[1] 
			\REQUIRE data matrix $\bm{X}$, (i.e., $\bm{H}^0$), adjacent matrix $\bm A$, one-hot label matrix $\bm{Y}_l$, parameter $\lambda$.\\
			\ENSURE the $\bm{Y}_u$.\\
			\STATE Initialize $\bm{Y}_u$ with random pseudo label matrix;
			\STATE Calculate the degree matrix $\bm{D}$, the Laplacian matrix $\bm{L}$ and the graph kernel  $\varphi(\bm A)$.
			\FOR{$a=0$ to $m$}
			\STATE Initialize $\bm{W}^{(a)}$ and $\bm{b}^{(a)}$;
			\ENDFOR 
			\WHILE{$not \; Convergence$}
			\FOR{$a=0$ to $m$}
			\STATE $\bm{H}^{(a)} \gets \sigma (\varphi(\widetilde{\bm A}) \bm{H}^{(a-1)} \bm{W}^{(a)} + \bm{b}^{(a)})$;
			\ENDFOR 
			\REPEAT
			\STATE Calculate $\bm{U}$ by Eq. (\ref{U_value});
			\STATE Update $\bm{Y}_u$ by Eq. (\ref{Y_u});
			\UNTIL{$Convergence$}
			\STATE Backward propagation with gradient descent;
			\ENDWHILE
		\end{algorithmic} 	
	\end{algorithm}
	
	\subsection{Time Complexity}
	In each iteration of the back propagation, the computaional complexity of graph convolution in Eq. (\ref{proposed model}) is $O(|\mathcal{E}|d_a)$, where $\bm{A}$ is a sparse matrix and the amount of non-zero entries is $|\mathcal{E}|$. Assume that the proposed deep graph model converges after $t$ iterations. The time complexity of a non-gradient decision layer is $O(u \log u+u^2c+ud_mc)$, where $u$ is unlabel nodes. Therefore, the total complexity is $O(|\mathcal{E}|d_a + t(ulogu+u^2c+ud_mc))$. Considering that the solution of the non-gradient decision layer is independent with gradient and there is no need to optimize  the classifier every iteration, the complexity can be approximate  to $O(|\mathcal{E}|d_a)$.

	\section{Proof} \label{section_opt}
	In this section, we will prove the proposed Theorem \ref{cal_U} and Theorem \ref{cal_Y_u} individually.
	
	\subsection{Proof of Theorem \ref{cal_U}}
	Since problem $\min \limits_{\bm{U}^T \bm{U}=I} ||\bm{X}\bm{U}-\bm{Y}||_F^2$ is difficult to solve directly, we introduce the matrix $\bm V \in R^{d \times (d-c)}$ generated in the orthogonal complement spaces of $\bm{U}$ according to \cite{2017Feature}. The problem is equivalent to the following balanced problem
	\begin{equation}
		\label{U_obj}
		\min \limits_{\bm{U}^T \bm{U}=\bm I} ||\bm{X}[\bm{U},\bm{V}]-[\bm{Y}, \bm{X}\bm{V}]||_F^2.
	\end{equation}
	
	To express more conveniently, we define the matrix $\bm{S}=[\bm{U}, \bm{V}]$ and problem can be transformed into
	\begin{equation}
		\label{U_obj_1}
		\begin{split}
			\min \limits_{\bm{S}^T\bm{S}=\bm{I}} &||\bm{X}\bm{S}-\bm{R}||_F^2 \\
			=\min \limits_{\bm{S}^T\bm{S}=\bm{I}} &Tr((\bm{X}\bm{S}-\bm{R})^T(\bm{X}\bm{S}-\bm{R}))\\
			=\min \limits_{\bm{S}^T\bm{S}=\bm{I}} &Tr({\bm{X}}^T\bm{S}^T \bm{S\bm{X}})-2Tr(\bm{S}^T\bm{X}\bm{R}),
		\end{split}
	\end{equation}
	where $\bm{R}=[\bm{Y}, \bm{X}\bm{V}]$.
	
	Owing to the orthogonal embedding, the problem can be simplified into
	\begin{equation}
		\label{U_obj_2}
		\min \limits_{\bm{S}^T\bm{S}=\bm{I}}  Tr({\bm{X}}^T\bm{X})-2Tr(\bm{S}^T\bm{X}\bm{R}).
	\end{equation}
	
	Notice that the first term in Eq. (\ref{U_obj_2}) is equal to a constant under the constraints. Therefore, the problem can be transformed into 
	\begin{equation}
		\label{U_solve}
		\max \limits_{\bm{S}^T\bm{S}=\bm{I}} Tr(\bm{S}^T\bm{X}\bm{R}).
	\end{equation}
	
	The following lemma \cite{9134971} points out that the transformed problem Eq. (\ref{U_solve}) has a closed-form solution.
	
	\begin{myLemma} 
		\label{Lemma_Q}
		For $\bm{Q}$, $\bm{P} \in {R}^{m \times n}$ where $m > n$, the problem 
		\begin{equation}
			\max \limits_{\bm{Q}^T \bm{Q} = \bm{I}} Tr(\bm{Q}^T \bm{P})
		\end{equation}
		can be solved by $\bm{Q} = \bm{B} \bm{C}^T$
		where $[\bm B, \bm \Psi, \bm{C}^T]=svd(\bm{P})$, $\bm B \in R^{m \times r}$, $\bm C \in R^{n \times r}$, and $r = rank(\bm P)$ . 
	\end{myLemma}
	
	According to Lemma \ref{Lemma_Q}, $\bm{S}$ can be calculated by
	\begin{equation}
		\label{L_value}
		\bm{S} = \bm{E} \bm{F}^T,
	\end{equation}
	where $[\bm E, \sim, \bm{F}^T]=svd(\bm{H}^{(m)}\bm{R})$. Therefore, the orthogonal matrix is obtained
	\begin{equation}
		\label{U_value}
		\bm{U} = \bm{S}[:,:c],
	\end{equation}
	where $\bm{S}[:,:c]$ means that the first $c$ columns of all the rows of $\bm S$ is taken.
	
	\begin{table*}[t]
		\vspace{-3mm}
		\centering
		\renewcommand\arraystretch{1.3}
		\caption{Datasets Description}
		\vspace{-4mm}
		\begin{tabular}{p{2.5cm}<{\centering}p{2.5cm}<{\centering}p{1.5cm}<{\centering}p{1.5cm}<{\centering}p{1.5cm}<{\centering}p{1.5cm}<{\centering}p{1.5cm}<{\centering}}
			\toprule
			\textbf{Dataset}	& \textbf{Type} 		& \textbf{Nodes} & \textbf{Edges} & \textbf{Features} & \textbf{Classes} & \textbf{Label Rate} \\ \hline
			Cora             	& Citation network		& 2$,$708           & 5$,$429           & 1$,$433              & 7                & 5.2\%               \\
			Citeseer         	& Citation network		& 3$,$327           & 4$,$732           & 3$,$703              & 6                & 3.6\%               \\
			Pubmed           	& Citation network		& 19$,$717          & 44$,$338          & 500               & 3                & 0.3\%               \\ 
			Coauthor-CS      	& Coauthor network		& 18$,$333          & 81$,$894          & 6$,$805              & 15               & 1.6\%               \\ 
			Coauthor-Phy     	& Coauthor network		& 34$,$496          & 247$,$962         & 8$,$415              & 5                & 57.9\%               \\ 
			\bottomrule
		\end{tabular}
		\label{table_datasets}
		\vspace{-4mm}
	\end{table*}

	\subsection{Proof of Theorem \ref{cal_Y_u}}
	It is difficult to optimize the second term in problem (\ref{Y_u_pro}) directly. Therefore, we transform it like 
	\begin{equation}
		\label{trans_y_u}
		\begin{split}
			min_{\bm Y_u}&\sum_{i,j=1}^n a_{ij}  \|\bm y_i - \bm y_j\|_2^2 \\
			= & \sum_{i,j=1}^n a_{ij} \|\bm y_i\|_2^2 + 					\sum_{i,j=1}^n a_{ij} \|\bm y_j\|_2^2 - 2\sum_{i,j=1}^n a_{ij} y_{ij}^2.\\
		\end{split}
	\end{equation}
	Considering that $\bm D=diag(\sum_j a_{ij})$, the Eq. (\ref{trans_y_u}) is formulated as
	\begin{equation}
		\label{trans_y_u_1}
		\begin{split}
			& \sum_{i,j=1}^n a_{ij} \|\bm y_i\|_2^2 + \sum_{i,j=1}^n a_{ij} \|\bm y_j\|_2^2 - 2\sum_{i,j=1}^n a_{ij} y_{ij}^2.\\
			=& \sum_{i=1}^n d_i \|\bm y_i\|_2^2 + \sum_{j=1}^n d_j \|\bm y_j\|_2^2 - 2\sum_{i,j=1}^n a_{ij} y_{ij}^2\\
			=& 2(\sum_{i=1}^n d_i \|\bm y_i\|_2^2-\sum_{i,j=1}^n a_{ij} y_{ij}^2)\\
			=& Tr(\bm Y^T \bm D \bm Y) - Tr(\bm Y^T \bm A \bm Y)\\
			=& Tr([\bm Y_l^T, \bm Y_u^T] \bm L [\bm Y_l; \bm Y_u]), \\
		\end{split}
	\end{equation}
	where $d_i$ represents the diagonal element in $\bm D$. Furthermore, the inital problem is written as
	%
	%
	%
	\begin{equation}
		\label{Y_u_optimze}
		\begin{split}
			\min \limits_{\bm{Y}_u} \mathcal{J}= &\underbrace{||[\bm{X}_l;\bm{X}_u]\bm{U} - [\bm{Y}_{l};\bm{Y}_{u}]||_F^2}_{\mathcal{J}_1} \\
			&+ \underbrace{\lambda Tr([\bm{Y}_{l}^T,\bm{Y}_{u}^T]
				[\begin{array}{cc}
					\bm{L}_{ll} & \bm{L}_{lu} \\
					\bm{L}_{ul} & \bm{L}_{uu}
				\end{array}]
				[\bm{Y}_{l};\bm{Y}_{u}])}_{\mathcal{J}_2},
		\end{split}
	\end{equation}
	where the $\bm{X}_l$ and $\bm{X}_u$ are the label and unlabel graph feature, respectively.

	The first parts in Eq. (\ref{Y_u_optimze}) can be transformed by
	\begin{equation}
		\label{J_1_trans}
		\begin{split}
			\mathcal{J}_1
			&=||[\bm{X}_l;\bm{X}_u]\bm{U} - [\bm{Y}_{l};\bm{Y}_{u}]||_F^2\\
			&=Tr(\bm{U}^T(\bm{X}_l^T \bm{X}_l+\bm{X}_u^T \bm{X}_u)\bm{U})
			+Tr(\bm{Y}_l^T\bm{Y}_{l}+\bm{Y}_u^T\bm{Y}_{u})\\
			&~~~~-2Tr(\bm{U}^T(\bm{X}_l^T\bm{Y}_{l}+\bm{X}_u^T\bm{Y}_{u}))\\
			&\Leftrightarrow Tr(\bm{Y}_u^T\bm{Y}_{u}) - 2Tr(\bm{U}^T\bm{X}_u^T\bm{Y}_{u}).
		\end{split}
	\end{equation}
	
	
	And the $\mathcal{J}_2$ can also be simplified by
	\begin{equation}
		\label{J_2_trans}
		\begin{split}
			\mathcal{J}_2
			&=\lambda Tr([\bm{Y}_{l}^T,\bm{Y}_{u}^T]
			[\begin{array}{cc}
				\bm{L}_{ll} & \bm{L}_{lu} \\
				\bm{L}_{ul} & \bm{L}_{uu}
			\end{array}]
			[\bm{Y}_{l};\bm{Y}_{u}])\\
			&=\lambda Tr(
			(\bm{Y}_{l}^T\bm{L}_{ll}+\bm{Y}_{u}^T\bm{L}_{ul})\bm{Y}_{l}
			+
			(\bm{Y}_{l}^T\bm{L}_{lu}+\bm{Y}_{u}^T\bm{L}_{uu})\bm{Y}_{u}
			)\\
			&\Leftrightarrow \lambda Tr(\bm{Y}_{u}^T\bm{L}_{ul}\bm{Y}_{l}
			+
			(\bm{Y}_{l}^T\bm{L}_{lu}+\bm{Y}_{u}^T\bm{L}_{uu})\bm{Y}_{u}
			)\\
			&\Leftrightarrow \lambda Tr(\bm{Y}_{u}^T\bm{L}_{uu}\bm{Y}_{u})+2\lambda Tr(\bm{Y}_{u}^T\bm{L}_{ul}\bm{Y}_{l}),
		\end{split}
	\end{equation}
	%
	with $\bm{L}_{lu}=\bm{L}_{ul}^T$.
	
	Combing the sub-problems Eq. (\ref{J_1_trans}) and Eq. (\ref{J_2_trans}), the primal problem can be define as
	\begin{equation}
		\label{transformed_J}
		\begin{split}
			\mathcal{J}
			=&Tr(\bm{Y}_u^T\bm{Y}_{u}) - 2tr(\bm{U}^T\bm{X}_u^T\bm{Y}_{u})\\
			&+\lambda Tr(\bm{Y}_{u}^T\bm{L}_{uu}\bm{Y}_{u})+2\lambda Tr(\bm{Y}_{u}^T\bm{L}_{ul}\bm{Y}_{l}).
		\end{split}
	\end{equation}
	
	Owing to a constraint on $\bm{Y}_u$, the problem Eq. (\ref{transformed_J}) is derivated w.r.t $\bm{Y}_u$ and set to 0. Then, we have
	\begin{equation}
		\label{Y_u}
		\begin{split}
			\bm{Y}_u
			=(\bm I + \lambda \bm{L}_{uu})^{-1}(\bm{X}_u\bm{U}-\lambda \bm{L}_{ul}\bm{Y}_l).
		\end{split}
	\end{equation}
	
	\subsection{Proof of Lemma \ref{Lemma_Q}}
	Having be decomposed with SVD, we can obtain the $[\bm B, \bm \Psi, \bm C^T]=svd(\bm{P})$. Note that
	\begin{equation}
		\begin{split}
			Tr(\bm Q^T \bm P)&=Tr(\bm Q^T \bm B \bm \Psi \bm C^T)=Tr(\bm C^T \bm Q^T \bm B \bm \Psi) \\
			&=Tr(\bm K \bm \Psi)=\sum_{i=1}^{n} k_{ii} \psi_{ii}
		\end{split}
		.
	\end{equation}
	where $\bm K = \bm C^T \bm Q^T \bm B$. Clearly, $\bm K \bm K^T = \bm I$ such that $k_{ij} \leq 1$. Hence, we have
	\begin{equation}
		Tr(\bm K \bm \Psi) \leq \sum_{i=1}^{n} \psi_{ii}.
	\end{equation}
	We can simply set $\bm K = \bm C^T \bm Q^T \bm B = \bm I_r $. In other words, 
	\begin{equation}
		\bm Q = \bm B \bm C^T.
	\end{equation}
	Consequently, the lemma is proved.
	

	\begin{figure*}[t]
		\centering
		\subfigure[Cora: 25 epoch]{
			\label{Cora: 25 epoch}
			\includegraphics[scale=0.28]{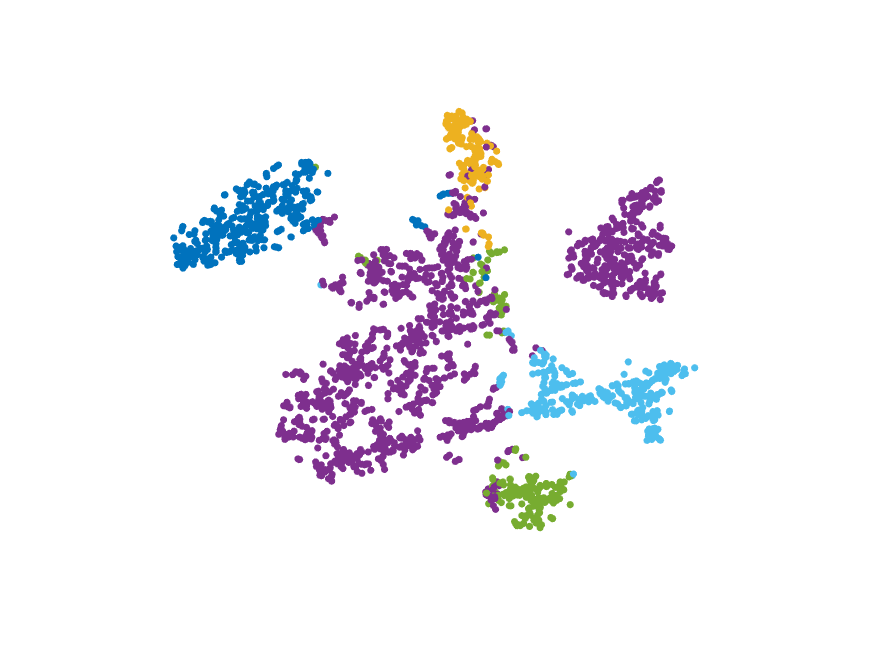}
		}
		\subfigure[Cora: 50 epoch]{
			\label{Cora: 50 epoch}
			\includegraphics[scale=0.28]{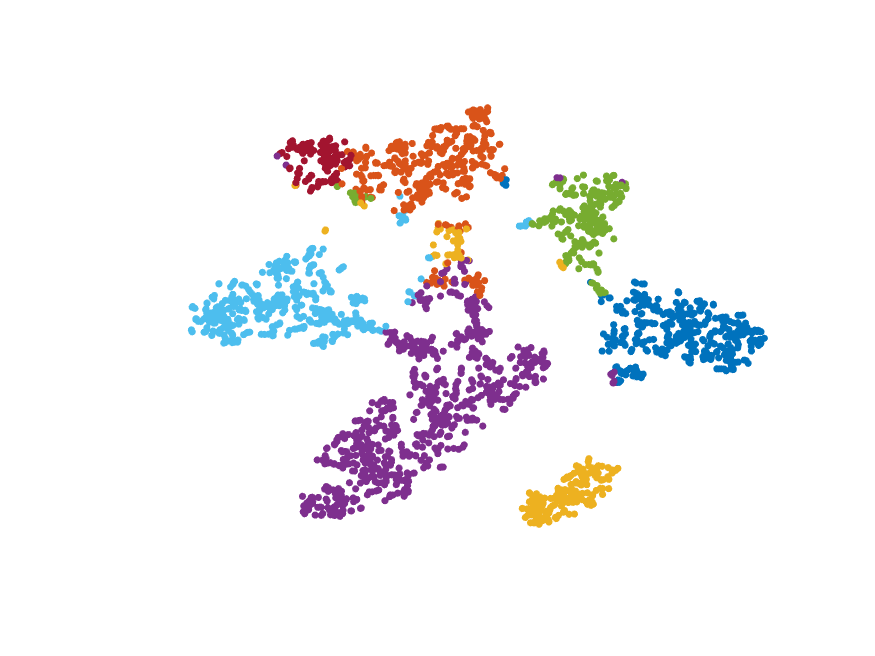}
		}
		\subfigure[Cora: 100 epoch]{
			\label{Cora: 100 epoch}
			\includegraphics[scale=0.28]{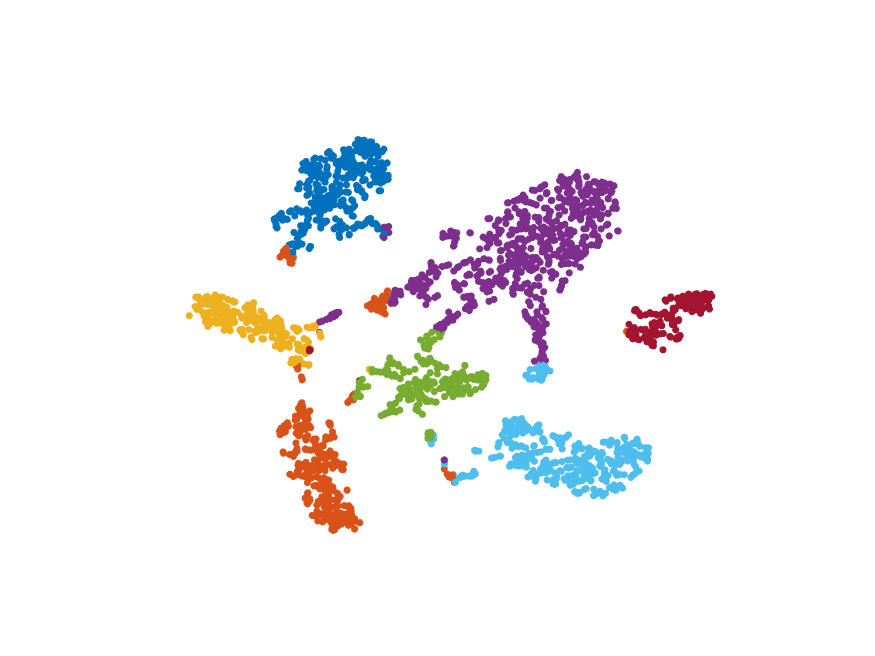}
		}
		\subfigure[Cora: 200 epoch]{
			\label{Cora: 200 epoch}
			\includegraphics[scale=0.28]{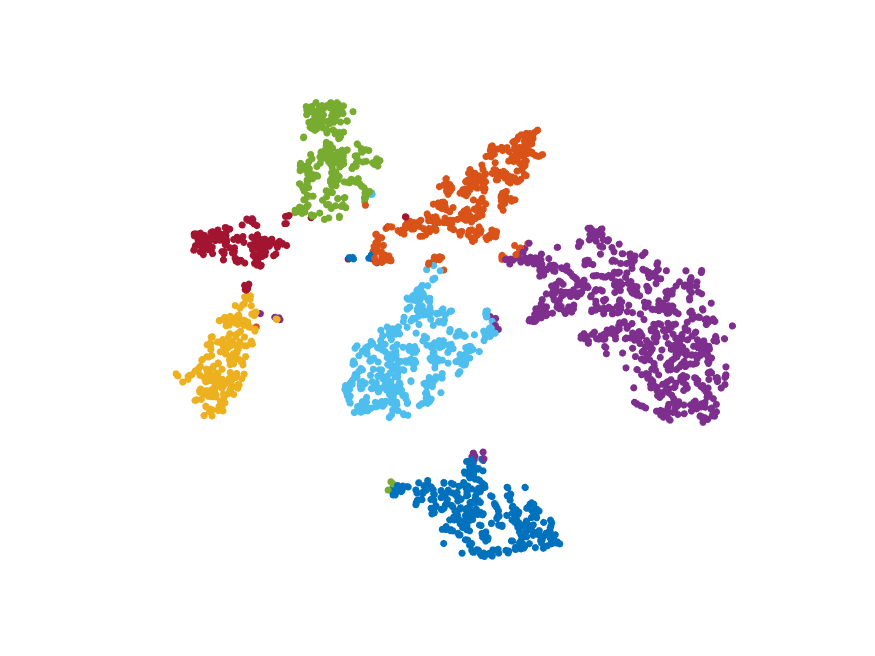}
		}
		\subfigure[Citeseer: 5 epoch]{
			\label{Citeseer: 5 epoch}
			\includegraphics[scale=0.28]{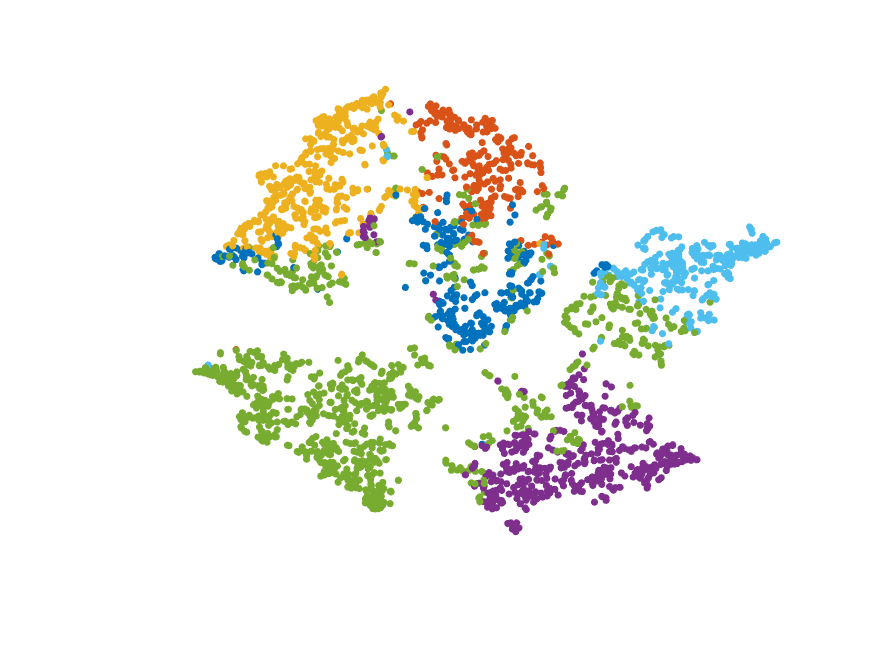}
		}
		\subfigure[Citeseer: 10 epoch]{
			\label{Citeseer: 10 epoch}
			\includegraphics[scale=0.28]{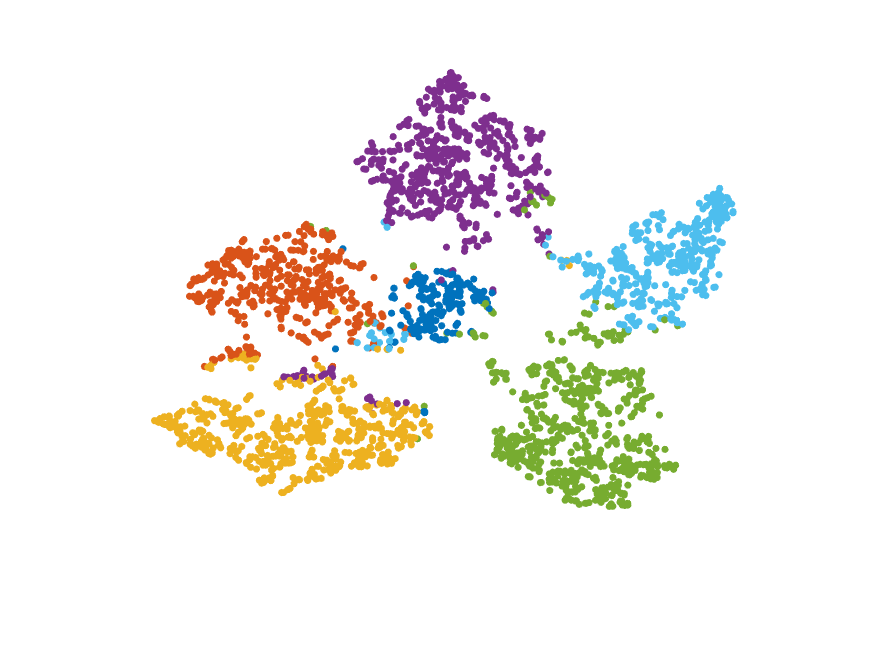}
		}
		\subfigure[Citeseer: 25 epoch]{
			\label{Citeseer: 25 epoch}
			\includegraphics[scale=0.28]{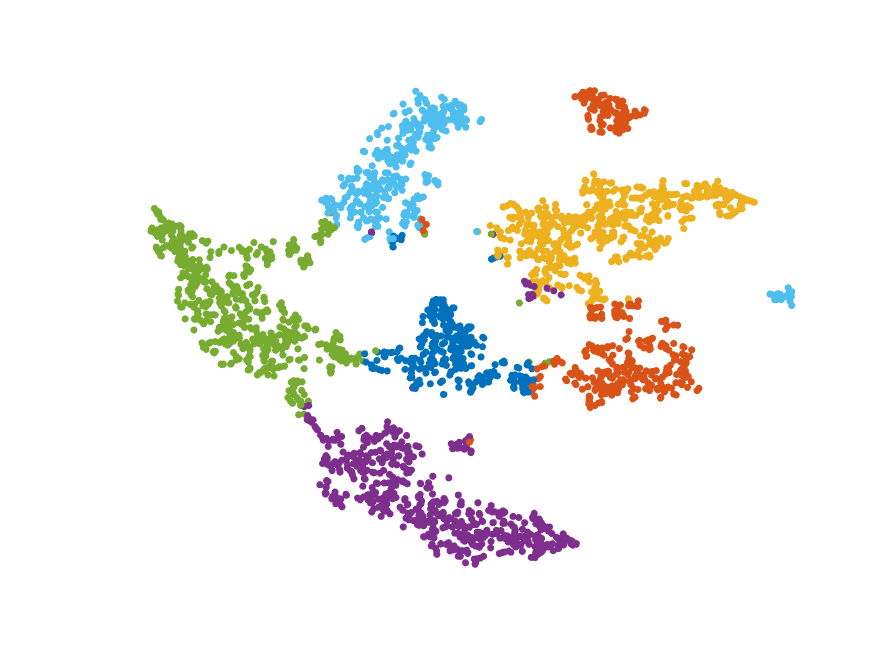}
		}
		\subfigure[Citeseer: 100 epoch]{
			\label{Citeseer: 100 epoch}
			\includegraphics[scale=0.28]{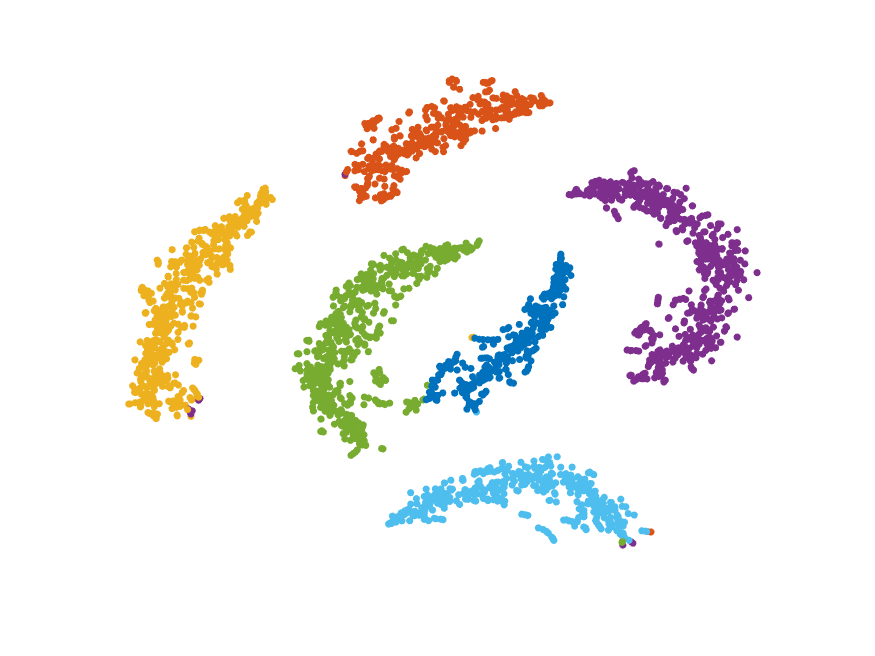}
		}
		\caption{Visualize the semi-supervised classification result via t-SNE \cite{2008Visualizing}. The colors represent the different classes. The above row \subref{Cora: 25 epoch} - \subref{Cora: 200 epoch}visualizes  the result of classification on the Cora dataset according to 25, 50, 100, and 200 epoch respectively. And the bottom row \subref{Citeseer: 5 epoch}- \subref{Citeseer: 100 epoch} shows the results on the Citeseer dataset according to 2, 10, 25, and 100 epoch respectively. }
		\label{Visualize the result}
	\end{figure*} 
	
	\begin{table*}[t]
		\vspace{-3mm}
		\centering
		\renewcommand\arraystretch{1.3}
		\caption{Semi-supervised classification accuracy (\%) on the benchmark dataset.}
		\vspace{-4mm}
		\begin{tabular}{p{3cm}p{1.5cm}p{2cm}<{\centering}p{2cm}<{\centering}p{2cm}<{\centering}p{2cm}<{\centering}p{2cm}<{\centering}}
			\toprule
			& \textbf{Input}		& \textbf{CORA}    		& \textbf{CITESEER} 	& \textbf{PUBMED}  	& \textbf{Coauthor-CS}		& \textbf{Coauthor-Phy}\\ 
			\hline
			\textbf{DeepEmbedding}  & $\bm{X}$,$\bm{Y}_l$,$\widetilde{\bm A}$	& 59.00 $\pm$ 1.72       		& 59.60 $\pm$ 2.29         	& 71.10 $\pm$ 1.37       & 70.93 $\pm$ 0.55			& 85.77 $\pm$ 0.08 \\ 
			\textbf{GAE}            & $\bm{X}$,$\bm{Y}_l$,$\widetilde{\bm A}$	& 71.50 $\pm$ 0.02   		& 65.80  $\pm$ 0.02          	& 72.10  $\pm$ 0.01				& 56.84 $\pm$ 0.43		& 66.20 $\pm$ 0.20 \\
			\textbf{GraphEmbedding} & $\bm{X}$,$\bm{Y}_l$	& 75.70 $\pm$ 3.83          		& 64.70 $\pm$ 1.11     	& 77.20 $\pm$ 4.38		& 28.91 $\pm$ 0.02		& 50.32 $\pm$ 1.11 \\
			\textbf{GCN}            & $\bm{X}$,$\bm{Y}_l$,$\widetilde{\bm A}$	& 81.50 $\pm$ 0.50          		& 70.30 $\pm$ 0.50          	& 78.33 $\pm$ 0.70		& 78.12 $\pm$ 3.42		& 92.64 $\pm$ 3.80 \\
			\textbf{DGI}            & $\bm{X}$,$\bm{Y}_l$	& 82.30 $\pm$ 0.60          		& 71.80 $\pm$ 0.70          	& 76.80 $\pm$ 0.60		& 84.34 $\pm$ 0.09		& 92.54 $\pm$ 0.03 \\
			\textbf{PFS-LS}			& $\bm{X}$,$\bm{Y}_l$	& 70.09 $\pm$ 1.38		& 43.23 $\pm$ 1.02		& 74.42 $\pm$ 14.2		& 58.80 $\pm$ 14.91		& 91.16 $\pm$ 6.55 \\
			\textbf{APPNP}			& $\bm{X}$,$\bm{Y}_l$,$\widetilde{\bm A}$	& 85.09 $\pm$ 0.25				& {71.93 $\pm$ 2.00}				& 79.73 $\pm$ 0.31		& \underline{91.52 $\pm$ 0.13}		& 95.56 $\pm$ 0.48 \\
			\textbf{GCN-LPA}		& $\bm{X}$,$\bm{Y}_l$,$\widetilde{\bm A}$	& 77.28 $\pm$ 0.16				& 70.72 $\pm$ 0.16				& 73.63 $\pm$ 0.40		& 87.63 $\pm$ 0.38		& \underline{95.80 $\pm$ 0.03} \\
			\textbf{MulGraph}       & $\bm{X}$,$\bm{Y}_l$,$\widetilde{\bm A}$	& \textbf{86.80 $\pm$ 0.50} 		& \underline{73.30 $\pm$ 0.50}   & \underline{80.10 $\pm$ 0.70}		& 88.36 $\pm$ 0.09		& 94.30 $\pm$ 0.01 \\
			\textbf{Ours}       & $\bm{X}$,$\bm{Y}_l$,$\widetilde{\bm A}$	& \underline{85.65 $\pm$ 0.76}  & \textbf{74.80 $\pm$ 0.41}  	& \textbf{80.87 $\pm$ 0.17}		& \textbf{91.60 $\pm$ 0.23}		& \textbf{95.89 $\pm$ 0.09} \\
			\bottomrule
		\end{tabular}
		\label{Contrast_result}
	\end{table*}

	\section{Experiment}
	To verify the performance of the proposed model, in this section we will compare the proposed model with seven state-of-the-art methods on citation network benchmark datasets.
	
	\subsection{Benchmark Dataset Description}
	We employ citation networks and coauthor networks to evaluate the performance of models:
	
	\noindent\textbf{Citation Networks:} 
	The three classic citation networks involved in our experiments are Cora, Citeseer, and Pubmed, which are closely following the experimental setup in \cite{yang2016revisiting}. In these networks, the nodes are articles and the edge reflects the citations between these articles. 
	
	\noindent\textbf{Coauthor Networks:}
	The two co-authorship networks \cite{2018Pitfalls}, Coauthor-CS and Coauthor-Phy, are also used for evaluation. In these networks, the nodes are authors and the edge suggests that two co-authored a paper. These features represent the keywords of the author's papers. Besides, the coauthor networks are split according to \cite{wang2020unifying}.
	
	The detail of these datasets is summarized in TABLE \ref{table_datasets}.

	\subsection{Experiment Setting}\label{experiment_setting}
	The ratio of labeled data for training is divided according to TABLE \ref{table_datasets}. Like the \cite{yang2016revisiting}, we employ the extra 500 labeled nodes as a valid set to optimize the hyperparameters including the hidden units in each layer, weight coefficient $\lambda$ in Eq. (\ref{proposed model}), dropout, and learning rate.
	
	\begin{figure*}[t]
		\vspace{-3mm}
		\centering
		\subfigure[Cora Convergence]{
			\label{Cora Convergence}
			\includegraphics[scale=0.30]{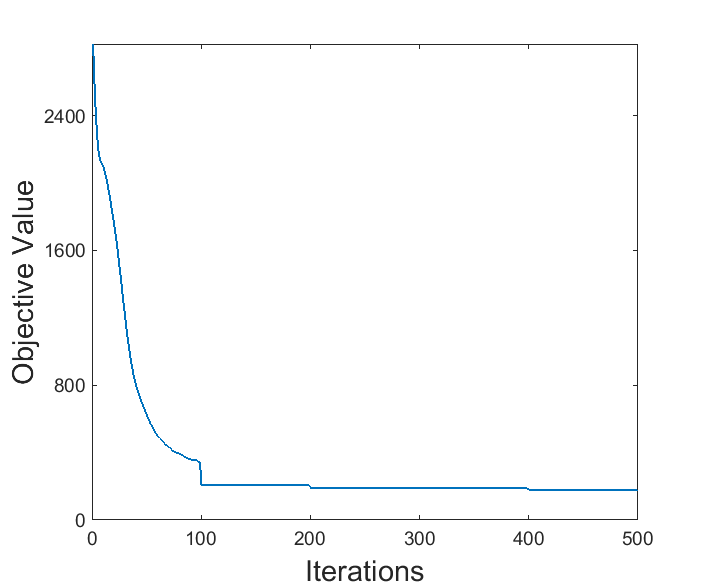}
		}
		\subfigure[Citeseer Convergence]{
			\label{Citeseer Convergence}
			\includegraphics[scale=0.30]{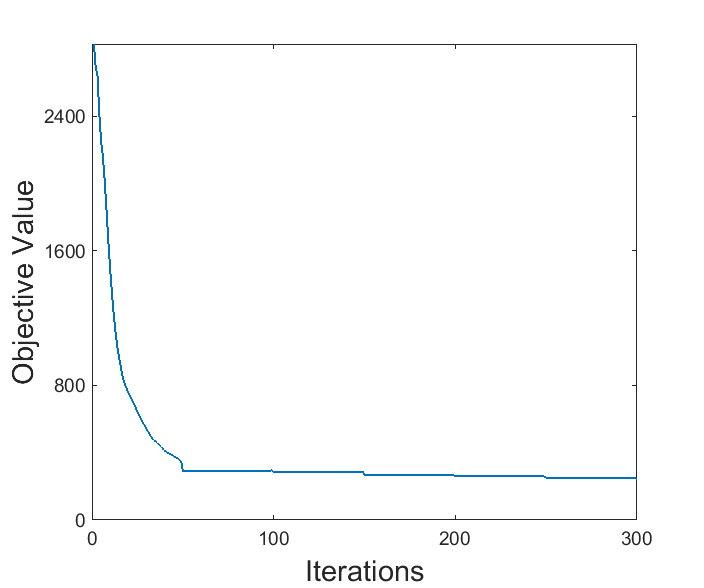}
		}
		\subfigure[Pubmed Convergence]{
			\label{Pubmed Convergence}
			\includegraphics[scale=0.30]{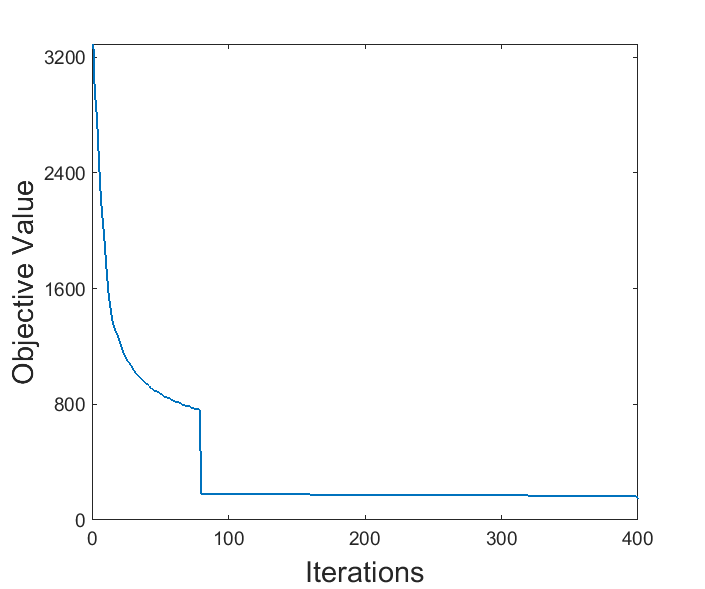}
		}
		\caption{The results of convergence on benchmark datasets including Cora, Citeseer, Pubmed. The objective value is defined in Eq. (\ref{proposed model}). And we choose every 100, 50, and 80 iterations to optimize the classifier on each dataset respectively. }
		\label{Convergence on benchmark}
	\end{figure*}

	\begin{figure}[t]
		\vspace{-6mm}
		\centering
		\subfigure[Convergence Epoch]{
			\includegraphics[scale=0.25]{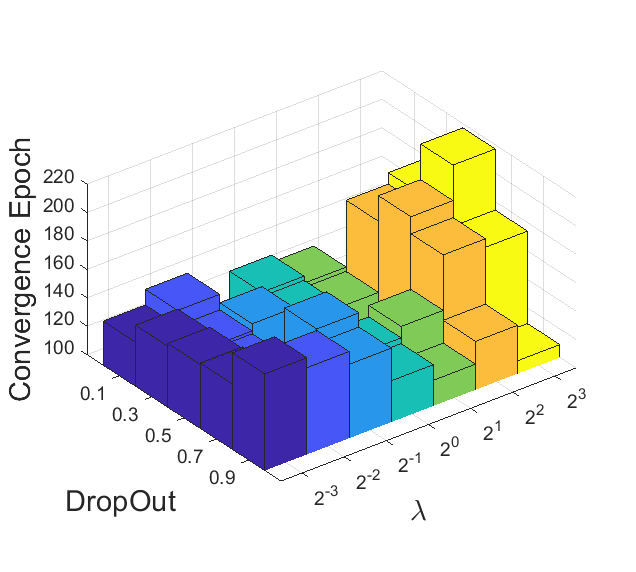}
			\label{Convergence Epoch}
		}
		\subfigure[Impact on Accuracy]{
			\includegraphics[scale=0.25]{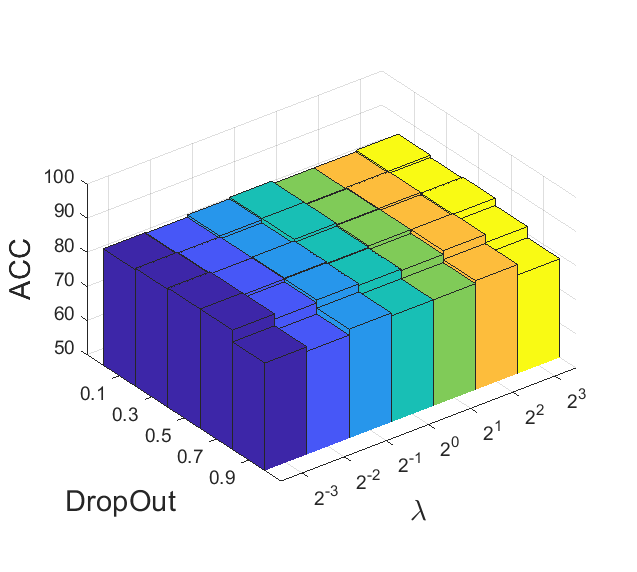}
			\label{ACC tune}
		}
		\caption{The sensitivity analysis of the parameters including dropout and coefficient weight $\lambda$ in Eq. (\ref{proposed model}) on Cora. \subref{Convergence Epoch} shows how dropout and $\lambda$ impact on the Convergence Epoch which means the epoch when the model has converged. And \subref{ACC tune} shows the influence of dropout and $\lambda$ on accuracy.
		}
		\vspace{-3mm}
		\label{Analysis of parameters}
	\end{figure}
	
	Before classification, the adjacent and feature matrix are accordingly (row-)normalized to the range of $[0, 1]$. The two graph convolution layers are used as a deep graph feature extractor. For the hidden units in each layer, dropout rate, learning rate, and weight coefficient $\lambda$, we choose values of parameters from $\left\{10, 32, 64, 128, 256, 512\right\}$, $\left\{0.1, 0.3, 0.5, 0.7, 0.9\right\}$, $\left\{10^{-3}, 10^{-2}, 10^{-1}, ..., 10^{2}, 10^{3}\right\}$ and $\left\{2^{-3}, 2^{-2}, 2^{-1}, ..., 2^{2}, 2^{3}\right\}$ respectively via 4-fold cross-validation and record the best records. The initialization described in \cite{glorot2010understanding} is employed to initialize the network weights. We train the model for a maximum of 500 training iterations with batch gradient descent. After tuning the hyperparameters, we adopt the adam optimizer with a learning rate of 0.01 and optimize the orthogonal manifold classifier every 50 iterations. Accuracy is employed to evaluate the performance of the model.

	\subsection{Comparsion with Methods} 
	We compare the proposed model with eight state-of-the-art semi-supervised models including Graph neural networks with personalized pagerank ($APPNP$) \cite{klicpera2019predict}, Unifying GCN and label propagation ($GCN$-$LPA$) \cite{wang2020unifying}, Multi-graph representation learning ($MulGraph$) \cite{hassani2020contrastive}, Deep graph infomax ($DGI$) \cite{velickovic2019deep},Parameter-Free Similarity of Label and Side Information ($PFS$-$LS$) \cite{8399869}, Graph Convolution Network ($GCN$) \cite{kipf2016semi}, Variational Graph Auto-Encoder ($GAE$) \cite{kipf2016variational}, Semi-supervised learning with Graph Embedding ($Graph Embedding$)\cite{yang2016revisiting} and Deep learning via semi-supervised  embedding ($DeepEmbedding$) \cite{weston2012deep}. Among them, $PFS$-$LS$ is a non-GNN model.
	
	To be fair, we utilize the same experiment setting in section \ref{experiment_setting}  train these comparison models and tune the correlated hyper-parameters. For classification models based graph, we utilize the self-loop adjacent $\widetilde{\bm A}$ to train as the same as in Eq. [\ref{proposed model}].  The models mentioned above are evaluated 10 times on the benchmark datasets. The accuracy and standard deviation are reported in TABLE \ref{Contrast_result}. Therefore, we could conclude that:
	\begin{itemize}
		\item [1)] The proposed model has achieved state-of-the-art accuracy with respect to the comparison methods. For example, on Citeseer and Pubmed dataset, we achieve $74.8\%$ and $80.87\%$ accuracy respectively. The scores are higher than the other methods, which indicates that the proposed model could mine the conceal relationship among points and work well on classification tasks with few labeled data. 
		\item [2)] When dealing with the datasets with high dimensions, huge points, and low labeling ratios like Pubmed and Coauthor-CS, it has a $1\%$ relative improvement over previous state-of-the-art. Besides, the proposed model has a good performance on Coauthor-Phy which has nearly 2.5 million edges and complex topological relationships.
		\item [3)] The proposed model is more successful and reasonable to integrate the graph topological structure with the label information than some state-of-the-art models like GCN-LPA, APPNP, and MulGraph. Among them, GCN-LPA utilizes label propagation to learn the edge weights of graphs. Compared with these models, ours integrates manifold learning with label local-structure as a non-gradient classifier. In this way, the deep graph networks could extract meaningful features to contribute to the decision layer.
		
	\end{itemize}
	
	The proposed model and comparison models are all implemented with PyTorch 1.2.0 on Windows 10 PC.
	
	\subsection{Convergence \& Visualization}
	Fig. \ref{Convergence on benchmark} shows the results on each benchmark datasets. It suggests that the proposed model converges rapidly on each dataset with $\lambda = 1$ during the training step. The decision layer (defined in Eq. (\ref{manifold_lp_loss})) is optimized after every 100, 50, and 80 backpropagation on each dataset respectively, which contributes to the objective value dropped sharply and accelerates the training of the proposed model. Besides, we investigate the impacts of the dropout in graph network and parameter $\lambda$ (defined in Eq. (\ref{proposed model})) on the model. The dropout ratio and $\lambda$ value are chosen like in Sec. \ref{experiment_setting}. 
	Convergence Epoch and Accuracy are utilized to evaluate the training speed and classification performance of the model. 
	The result is shown in Fig. \ref{Analysis of parameters}. From the results,  we notice that the Convergence Epoch varies considerably with the $\lambda$ changing. On the contrary, Accuracy is not sensitive to weight value. 
	In conclusion, the $\lambda$ mainly has an impact on the convergence rate of the model, and the proper dropout ratio could enhance the performance of the proposed model via discarding some neurons during the training.
	
	Apart from that, we utilize the t-SNE \cite{2008Visualizing} to reduce the deep graph embedding $\bm{H}^{(m)}$ (defined in Eq. (\ref{proposed model})) to 2D. Before visualize, these embeddings are not normalized and the dimension is not reduced via PCA and normalized. The perplexity of t-SNE is set to 30.  And the visualization of the reduction results in Fig. \ref{Visualize the result}. We notice that the topological data is not preserved well at the beginning of the training. When the algorithm has converged,  the distribution of the deep graph embedding is evenly and could be easy to classify.
	
	\begin{table}[t]
		\centering
		\vspace{-3mm}
		\renewcommand\arraystretch{1.3}
		\caption{Ablation Study on Benchmark Dataset.}
		\vspace{-4mm}
		\begin{tabular}{cccc}
			\toprule
			& \textbf{Cora}  		& \textbf{Citeseer}			& \textbf{Pubmed}  \\ \hline
			\textbf{GCN-Softmax}	& 78.99 $\pm$ 0.61      & 70.39 $\pm$ 0.58			& 78.39 $\pm$ 0.61 \\
			\textbf{GCN-LP}			& 79.70	$\pm$ 0.52		& 71.52 $\pm$ 0.28			& 78.67 $\pm$ 0.47 \\
			\textbf{GCN-OM}     	& \underline{83.91 $\pm$ 0.23}   	& \underline{73.71 $\pm$ 0.16}			& \underline{79.82 $\pm$ 0.72} \\
			\textbf{OURS}   	& \textbf{85.65 $\pm$ 0.76}		& \textbf{74.80 $\pm$ 0.41}		    & \textbf{80.87 $\pm$ 0.17} \\ 
			\bottomrule
		\end{tabular}
		\label{ablation_study}
		\vspace{-4mm}
	\end{table}
	
	\subsection{Ablation Study}
	We conduct an ablation study to evaluate how each part of the proposed non-gradient graph layer, Orthogonal Manifold (OM) and Local-structure Preservation (LP), contribute to overall model performances. Besides, GCN with a softmax classification strategy is employed as a baseline. The setting and results are summarized in Table \ref{ablation_study}.  
	We confirm that graph mining methods, orthogonal manifold learning and local-structure preservation, can both learn the relationship knowledge among the deep embedding. Besides, it is shown that the accuracy of GCN-OR and GCN-LP are both superior to traditional GCN-softmax. Moreover, the proposed deep model with a non-decision layer not only properly unifies the OM and LP parts but also successfully utilizes the topological structure and label information to enhance the performance.

	\section{Conclusion}
	In this paper, we propose a deep graph model with a non-gradient decision layer. 
	Unifying the orthogonal manifold and label local-structure preservation, the deep graph model successfully learns the topological knowledge of the deep embedding. Furthermore, compared with softmax optimized via gradient descent, the non-gradient layer can be solved with the analytical solutions via the proposed theorems. On the benchmark datasets, the proposed model can achieve higher accuracy over the previous state-of-the-art works.

	\bibliographystyle{IEEEbib.bst}
	\bibliography{mtl.bib}
	
\end{document}